\documentclass[10pt,letterpaper]{article}
\usepackage[margin=0.9in,headheight=14pt]{geometry}
\usepackage[T1]{fontenc}
\usepackage{lmodern}
\usepackage{amsmath,amssymb}
\usepackage{graphicx}
\usepackage{booktabs,array,tabularx,longtable,colortbl}
\usepackage{microtype}
\usepackage{fancyhdr}
\usepackage[numbers,sort&compress]{natbib}
\usepackage{url}
\usepackage[hidelinks]{hyperref}
\usepackage{xcolor}

\definecolor{AnluTableFirst}{HTML}{B22222}
\definecolor{AnluTableSecond}{HTML}{225EA8}
\definecolor{AnluTableThird}{HTML}{26734D}
\definecolor{AnluTableMuted}{HTML}{777777}
\definecolor{AnluTableHead}{HTML}{EDF1F5}
\definecolor{AnluTableOurs}{HTML}{FFF4D6}
\newcommand{\AnluTopOne}[1]{\textcolor{AnluTableFirst}{\textbf{#1}}}
\newcommand{\AnluTopTwo}[1]{\textcolor{AnluTableSecond}{\underline{#1}}}
\newcommand{\AnluTopThree}[1]{\textcolor{AnluTableThird}{\textit{#1}}}

\definecolor{AnluWordmarkGreen}{HTML}{24584D}
\newcommand{\anluwordmark}{\textcolor{AnluWordmarkGreen}{\textbf{A\kern0.02em n\kern0.02em l\kern0.02em u}}}

\newcommand{\rev}[1]{{\color{black}#1}}
\newcolumntype{Y}{>{\raggedright\arraybackslash}X}
\newcolumntype{L}[1]{>{\raggedright\arraybackslash}p{#1}}
\newcommand{\anlufigure}[2][\linewidth]{\includegraphics[clip,width=#1]{#2}}

\title{\anluwordmark: Enabling In-Context Time Series Anomaly Detection in Foundation Models via Counterfactual Supervision}
\author{%
  \normalsize Tian Lan$^{1}$ \quad Yifei Gao$^{1}$ \quad Yimeng Lu$^{1}$ \quad Xuming An$^{1}$\\[3pt]
  \normalsize Meng Wang$^{2}$ \quad Yue Pan$^{2}$ \quad Wenjun He$^{2}$ \quad
  Chen Zhang$^{1}$\thanks{Corresponding author: \href{mailto:zhangchen01@tsinghua.edu.cn}{\texttt{zhangchen01@tsinghua.edu.cn}}.}\\[6pt]
  \small $^{1}$Department of Industrial Engineering, Tsinghua University\\
  \small $^{2}$Huawei\\[5pt]
  \footnotesize\texttt{gao-yf@mail.tsinghua.edu.cn}\\
  \footnotesize\texttt{\{lant23,luym25,axm24\}@mails.tsinghua.edu.cn}\\
  \footnotesize\texttt{\{wangmeng71,panyue33,hewenjun8\}@huawei.com}}
\date{\small Preprint}
\hypersetup{
  pdftitle={Anlu: Enabling In-Context Time Series Anomaly Detection in Foundation Models via Counterfactual Supervision},
  pdfauthor={Tian Lan, Yifei Gao, Yimeng Lu, Xuming An, Meng Wang, Yue Pan, Wenjun He, Chen Zhang},
  pdfsubject={Preprint}}
\renewcommand{\arraystretch}{1.12}
\ifdefined\pdfminorversion\pdfminorversion=7\fi

\begin{document}
\raggedbottom
\maketitle

\begin{abstract}
\rev{Whether a time-series pattern is anomalous often depends on the operating regime of the monitored process. A missing event can signal a fault in one regime and be routine in another, and the query alone may not reveal which regime applies. We study in-context learning (ICL) for time series anomaly detection (TSAD) through reference-conditioned detection, where a reference record provides evidence about expected behavior and model parameters remain fixed at inference. Supplying the reference is not enough: when training anomalies are recognizable from the query alone, the detector can fit its targets while ignoring the reference. We therefore introduce counterfactual supervision, which pairs one query with two references that support different normal rules and labels the query under each. At positions where the two labels disagree, no detector that ignores the reference can fit both targets. Anlu learns from this supervision by adding a reference memory and zero-initialized gated adapters to a frozen time-series foundation model (TSFM) pretrained for anomaly detection. On the 350 TSB-AD-U evaluation sequences, Anlu raises the mean VUS-PR of the frozen TSFM from 0.542 to 0.607. Replacing the reference with zeros lowers Anlu's score to 0.499.}
\end{abstract}

\noindent\textit{Keywords:} In-context learning, Time series anomaly detection, Reference conditioning, Counterfactual supervision

\section{Introduction}
\label{sec:introduction}

\begin{figure}[!htbp]
  \centering
  \anlufigure[0.86\linewidth]{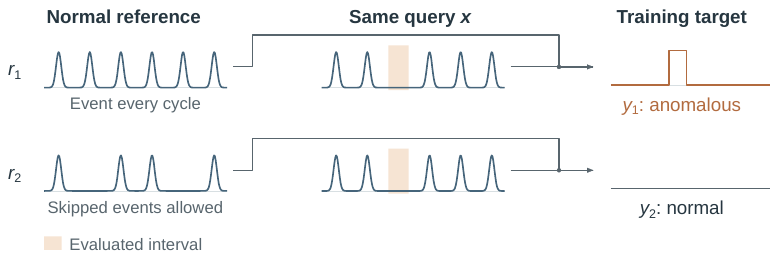}
  \caption{One fixed query under two normal references. The orange span is the evaluated query interval. Reference $r_1$ repeats the scheduled event on every cycle, so the missing query event is anomalous. Reference $r_2$ repeats admissible omissions, so the same interval is normal. These labels are the conditional training targets.}
  \label{fig:reference-conditioning}
\end{figure}

\rev{Industrial monitoring judges a sensor stream against the normal behavior of the process that produced it. That behavior changes with products, loads, and maintenance schedules, and a pattern that is normal in one operating regime can be a fault in another. A missing pulse indicates a fault when the process must emit an event on every cycle, but it is routine when intermittent events are allowed. The query alone may not reveal which rule applies. Figure~\ref{fig:reference-conditioning} keeps one query fixed and changes the normal reference against which it is judged.}

\rev{We frame this setting as reference-conditioned in-context learning: the supplied reference provides evidence about expected behavior for anomaly scoring, while model parameters remain fixed.}

\rev{Current detectors obtain regime information in one of two ways. Zero-shot detectors built on pretrained time-series foundation models (TSFMs) score a query from its own values and surrounding context \citep{goswami2024moment,lan2025towards}, so they cannot be told which regime applies. Semi-supervised detectors fit their parameters to normal data from the target series \citep{liu2024elephant}, so each new regime requires another round of training. Records of normal operation are easy to collect, whereas representative faults are rare. We study reference-conditioned detection, in which a separate normal record enters the detector as an input and specifies the regime at inference. Normal operation can then be demonstrated before any fault has been observed.}

\rev{Adding a reference input does not ensure that the detector uses it. When a synthetic anomaly is recognizable from the query alone, training can reach a low loss with one rule for every reference. Counterfactual supervision removes this shortcut. Each training unit pairs one query with two references that support different normal rules and labels the query under each, so that some positions are anomalous under one reference and normal under the other. A detector that ignores the reference assigns the same score to such a position in both branches, and its mean cross-entropy there cannot fall below $\log 2$ (Section~\ref{sec:counterfactual}). Fitting both labels therefore requires reading the reference.}

\rev{Anlu adds this reference pathway to a frozen TSFM pretrained for pointwise anomaly detection. A resampler compresses the reference into a fixed-length memory, and gated adapters read that memory at four depths of the query encoder. Only the resampler and the adapters are trained. Because the gates start at zero, Anlu reproduces the frozen TSFM at initialization.}

\rev{On the 350 TSB-AD-U evaluation sequences, Anlu raises the mean VUS-PR of the frozen TSFM from 0.542 to 0.607. To test whether this gain comes from the reference content, we keep the weights, queries, reference lengths, and validity masks fixed and set the normalized reference values to zero. The mean falls to 0.499, below the frozen TSFM, and the original reference scores higher on 248 of the 350 sequences.}

\rev{Our contributions are as follows.
\begin{itemize}
  \item We formulate reference-conditioned anomaly detection and introduce counterfactual supervision, in which one query receives conflicting labels under two references. At label-changing positions, any detector that ignores the reference incurs a cross-entropy of at least $\log 2$ (Section~\ref{sec:counterfactual}).
  \item We present Anlu, which adds a reference memory and zero-initialized gated adapters to a frozen TSFM and trains only this pathway (Section~\ref{sec:anlu}).
  \item On TSB-AD-U, Anlu raises the mean VUS-PR of the frozen TSFM from 0.542 to 0.607, and a zero-reference intervention lowers it to 0.499 (Section~\ref{sec:experiments}).
\end{itemize}}

\section{Related Work}
\label{sec:related}

\rev{\paragraph{Time-series foundation models.}
Pretrained TSFMs transfer temporal representations across datasets without per-dataset training. TimesFM, Chronos, Moirai, and Toto target zero-shot forecasting, and MOMENT covers forecasting, classification, imputation, and anomaly detection with one pretrained model \citep{das2023decoder,ansari2024chronos,woo2024unified,goswami2024moment,cohen2026time}. Zero-shot anomaly detectors pretrain on multi-domain or synthetic data and score a query window directly. DADA uses adaptive bottlenecks and dual adversarial decoders \citep{shentu2025towards}, and TimeRCD learns pointwise scores from relative context discrepancy \citep{lan2025towards}. These detectors infer normal behavior from the query window itself and accept no separate record of normal operation. TSFMs have begun to use series beyond the target: in-context fine-tuning prompts TimesFM with examples from related series \citep{das2024context}. The forecasting target is still the future of the target series, so the examples can improve a forecast but do not change what is forecast. In Anlu, the reference determines the target itself, and the same query receives a different label when the reference supports a different normal rule.}

\rev{\paragraph{In-context anomaly detection.}
In-context learning adapts a pretrained model to a task through examples in its input, with parameters fixed \citep{brown2020language}. Several anomaly detectors adopt this interface. InCTRL scores an image by its residuals to a few normal sample prompts \citep{zhu2024toward}. LLMAD retrieves similar normal and anomalous time-series segments as demonstrations for a large language model \citep{liu2025large}. iAmTime trains a TSFM on demonstration episodes that include corrupted series paired with anomaly masks, and it selects the anomaly type per episode so that the demonstrations indicate which patterns count as anomalous \citep{saha2026foundation}. In InCTRL and LLMAD, the label of a query is set by its own content, so the context can help detection without being necessary for it. iAmTime varies which anomaly types count across demonstrations. Anlu instead keeps the query fixed and varies the normal rule supported by a reference, and counterfactual supervision assigns labels that differ where the two references support different rules.}

\section{Method}
\label{sec:method}

\subsection{Reference-conditioned tasks}
\label{sec:counterfactual}
\label{sec:problem}

Let $x\in\mathbb R^{T\times C}$ be a query of length $T$ and $r\in\mathbb R^{L\times C}$ a reference of length $L$ over the same $C$ channels. Their lengths can differ and their samples need not be aligned. Given frozen parameters $\theta_0$ and adaptation parameters $\phi$, the detector returns a score for every query time,
\begin{equation}
 s=f_{\theta_0,\phi}(x,r)\in[0,1]^T.
 \label{eq:detector}
\end{equation}

For synthetic tasks, a normal rule defines admissible behavior, and the binary conditional label $y_r\in\{0,1\}^T$ of the query follows that rule and the evidence realized in reference $r$.

A counterfactual unit contains one query and two labeled references,
\begin{equation}
 \tau_i=(x_i,r_i^a,y_i^a,r_i^b,y_i^b),\qquad
 \mathcal D_i=\{t\in V_i^a\cap V_i^b:y_{i,t}^a\ne y_{i,t}^b\},
 \label{eq:tuple}
\end{equation}
where $V_i^a$ and $V_i^b$ are the valid labeled positions and $\mathcal D_i$ collects positions whose labels differ. The generator keeps units with $\mathcal D_i\ne\varnothing$ and with reference evidence for both labels. It varies transition speed, skipped states, event strength or absence, timing, and activity at scheduled idle slots. Appendix~\ref{app:generation} records the rule specification.

For example, the generator renders one query in which a scheduled event is suppressed. One reference repeats that event. The other repeats both occurrences and omissions. Absence of the event is anomalous under the first reference and normal under the second.

\rev{Conflicting labels exclude query-only solutions at label-changing positions. Suppose a deterministic detector ignores the reference and assigns the same probability $p\in(0,1)$ to a position $t\in\mathcal D_i$ in both branches. Its equally weighted cross-entropy at $t$ is
\begin{equation}
 -\tfrac12\log p-\tfrac12\log(1-p)\geq\log 2,
 \label{eq:query-only-bound}
\end{equation}
with equality at $p=1/2$. A loss below $\log 2$ at $t$ therefore requires the score to depend on the reference.}

\subsection{Reference memory and gated detection}
\label{sec:anlu}

Anlu initializes the query encoder and the reference encoder from the same frozen TSFM. The TSFM is a Toto-style network \citep{cohen2026time} pretrained on a separate synthetic univariate corpus with masked-query reconstruction and pointwise anomaly classification. Adaptation freezes both encoders, the query pointwise projection, and the detection and reconstruction heads.

For channel $c$, the reference encoder produces patch representations $R_c\in\mathbb R^{P_r\times d}$. A trainable Perceiver-style resampler\rev{, as in Flamingo \citep{alayrac2022flamingo},} maps them to a memory of $K$ tokens,
\begin{equation}
 M_c=\mathcal R_\phi(R_c;m_r)\in\mathbb R^{K\times d},
 \label{eq:memory}
\end{equation}
where $P_r$ is the number of reference patches, $d$ is the encoder width, and $m_r$ marks the unpadded reference patches. Each adapted query-encoder depth reads this memory with its own interaction parameters.

\begin{figure}[!htbp]
 \centering
 \anlufigure[\linewidth]{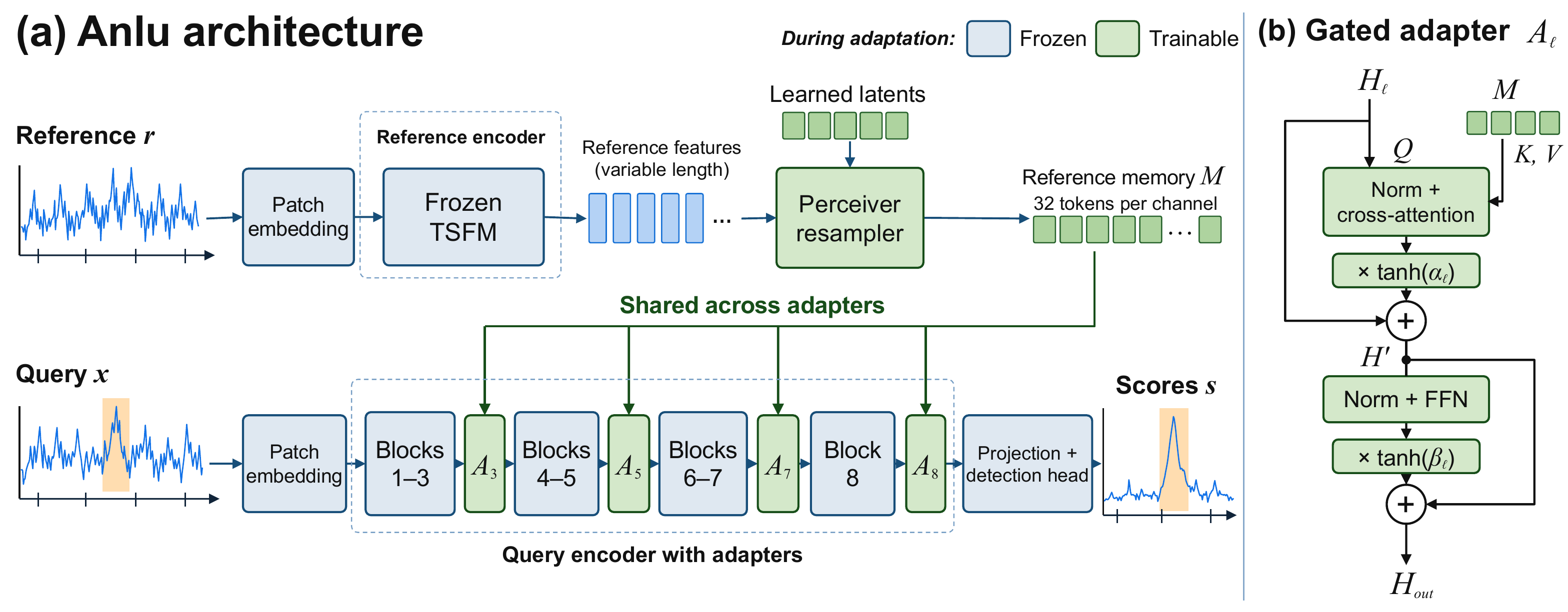}
 \caption{Anlu's framework. (a) The reference and query encoders use the same pretrained TSFM initialization. A Perceiver resampler compresses variable-length reference features into 32 memory tokens per channel. Adapters after query blocks 3, 5, 7, and 8 read the shared memory, and the frozen projection and detection head produce query-aligned scores. (b) Each adapter applies normalized cross-attention and a feed-forward block, each followed by a zero-initialized tanh gate and a residual connection. Blue denotes frozen modules and green denotes modules trained during adaptation.}
 \label{fig:anlu-architecture}
\end{figure}

Given query-patch representations $H_{\ell,c}$ after frozen block $\ell$, an adapter applies
\begin{equation}
\begin{aligned}
 H'_{\ell,c}&=H_{\ell,c}+\tanh(\alpha_\ell)A_\ell(H_{\ell,c},M_c),\\
 \widetilde H_{\ell,c}&=H'_{\ell,c}+\tanh(\beta_\ell)F_\ell(H'_{\ell,c}),
\end{aligned}
\label{eq:adapter}
\end{equation}
where $A_\ell$ is normalized cross-attention with an output projection and $F_\ell$ is a normalized feed-forward block. The scalar gates $\alpha_\ell$ and $\beta_\ell$ start at zero, so the inserted adapter is an identity map at initialization.

The detection head returns two logits per time and channel, $u_{t,c,1}$ and $u_{t,c,0}$. Their difference defines the anomaly logit and score,
\begin{equation}
 a_t=\frac1C\sum_{c=1}^{C}(u_{t,c,1}-u_{t,c,0}),\qquad
 s_t=\operatorname{sigmoid}(a_t).
 \label{eq:scores}
\end{equation}
The eight-block encoder has width 512 and patch size 16. A one-layer resampler produces 32 memory tokens per channel. Adapters follow blocks 3, 5, 7, and 8. Appendix~\ref{app:implementation} lists the full configuration.

\subsection{Preprocessing and training}
\label{sec:objective}
\label{sec:preprocessing}

The synthetic training set contains \rev{about one million} examples from reference-conditioned pairs and from anomaly-injection examples. For each channel, the mean $\mu_c$ and population standard deviation $\sigma_c$ of the full query are computed before cropping and applied to both series:
\begin{equation}
 \bar x_{t,c}=\frac{x_{t,c}-\mu_c}{\eta_c},\qquad
 \bar r_{t',c}=\frac{r_{t',c}-\mu_c}{\eta_c},\qquad
 \eta_c=\max(\sigma_c,10^{-4}).
 \label{eq:normalization}
\end{equation}
The shared query statistics preserve the relative scale of the two series. Here $t$ indexes the query and $t'$ indexes the reference.

Each stored example is cropped and masked on its own. A query crop contains at most 10,000 samples. If the query contains labeled anomalies, the crop includes a randomly chosen anomalous time with probability 0.5, and otherwise its start is uniform. A query with no labeled anomaly uses a uniform start. The reference crop is an independent contiguous interval of length 90\% to 100\% of the stored reference. Independent masks are applied to 15\% of the query patches and 15\% of the reference patches. Appendix~\ref{app:training} gives the composition of the training set.

For a minibatch $\mathcal B$, let $V_i$ be the valid labeled times of example $i$. The detection loss is the mean binary cross-entropy over those times,
\begin{equation}
 \mathcal L_{\mathrm{det}}=
 -\frac{\displaystyle\sum_{i\in\mathcal B}\sum_{t\in V_i}
 [y_{i,t}\log s_{i,t}+(1-y_{i,t})\log(1-s_{i,t})]}
 {\displaystyle\sum_{i\in\mathcal B}|V_i|}.
 \label{eq:detection-loss}
\end{equation}

The frozen reconstruction head predicts the normalized, unmasked query at the masked valid positions $\mathcal M$. The training objective is
\begin{equation}
 \mathcal L(\phi)=\mathcal L_{\mathrm{det}}(\phi)
 +\frac{1}{|\mathcal M|C}\sum_{(i,t)\in\mathcal M}\sum_{c=1}^{C}
 (\widehat x_{i,t,c}-\bar x_{i,t,c})^2.
 \label{eq:training-loss}
\end{equation}

AdamW trains $\phi$ at batch size 24, with a constant learning rate of $10^{-4}$ and weight decay $10^{-5}$.

\section{Empirical Evaluation}
\label{sec:experiments}

\subsection{Evaluation setup}
\label{sec:evaluation-setup}

We evaluate Anlu on the 350 sequences in TSB-AD-U Eval, from 23 source datasets \citep{liu2024elephant}. The metric is VUS-PR, the volume under the precision--recall surface \citep{paparrizos2022volume}, and the reported value is the unweighted mean across sequences.

TSB-AD specifies a training boundary $b$ for each sequence. The reference is the first $\min(b,10{,}000)$ samples after rows with missing values are removed. Annotated anomalies occur in 35 of the 350 prefixes, and the reported mean includes those sequences.

The query is normalized once by its mean and population standard deviation, and the reference uses those statistics. The zero-reference condition then sets every normalized reference value to 0 and keeps the reference length and validity mask. Long queries are scored in overlapping windows, and anomaly logits are averaged on the overlaps. At inference the random patch mask is off, and padding is handled by the validity mask. Appendix~\ref{app:evaluation} gives the window length, stride, and metric implementation.

\paragraph{Univariate benchmark comparison.}
Table~\ref{tab:univariate-complete} starts from the 700 series in the 11 univariate datasets of TimeRCD Table~1 \citep{lan2025towards}. We exclude a series if any point in its designated training prefix has an anomaly label, so that the prefix used as reference contains no annotated anomalies. This removes 168 series and leaves 532 series from 10 datasets. We recompute the dataset means for every baseline from its archived per-series results on this same subset, and evaluate Anlu with a window of 10,000 and a stride of 1,500.

\clearpage
\begingroup
\fontsize{8}{10}\selectfont
\setlength{\tabcolsep}{2.2pt}
\renewcommand{\arraystretch}{1.08}
\setlength{\LTcapwidth}{\textwidth}
\setlength{\LTleft}{\fill}
\setlength{\LTright}{\fill}
\begin{longtable}{@{}l*{10}{r}rccc@{}}
\caption{Univariate results after excluding series with anomalies in their training prefixes from the datasets in TimeRCD Table~1. The filter retains 532 series from 10 datasets; no Stock series remain. All models are aggregated over the same retained series. Each cell is the mean per-series score multiplied by 100. Ranks compare eligible models jointly across the zero-shot and full-shot groups.}\label{tab:univariate-complete}\\
\toprule
Model & IOPS & MGAB & NAB & NEK & Power & SED & TODS & UCR & WSD & YAHOO & \shortstack{Avg. rank\\$\downarrow$} & 1st & 2nd & 3rd \\
\midrule
Retained series & 16 & 9 & 26 & 5 & 1 & 3 & 2 & 228 & 105 & 137 & & & & \\
\midrule
\endfirsthead
\caption[]{Univariate results with anomaly-free training prefixes, continued.}\\
\toprule
Model & IOPS & MGAB & NAB & NEK & Power & SED & TODS & UCR & WSD & YAHOO & \shortstack{Avg. rank\\$\downarrow$} & 1st & 2nd & 3rd \\
\midrule
Retained series & 16 & 9 & 26 & 5 & 1 & 3 & 2 & 228 & 105 & 137 & & & & \\
\midrule
\endhead
\bottomrule
\multicolumn{15}{@{}p{\textwidth}@{}}{\vspace{3pt}\fontsize{7.1}{9}\selectfont \AnluTopOne{First}, \AnluTopTwo{second}, and \AnluTopThree{third} place are highlighted per dataset. $^\ast$ preserves the source table\textquotesingle s ranking exclusions for DADA and MOMENT. Excluded cells enter neither average ranks nor counts; a subscript gives the number of ranked datasets when fewer than 10.}\\
\endfoot
\bottomrule
\multicolumn{15}{@{}p{\textwidth}@{}}{\vspace{3pt}\fontsize{7.1}{9}\selectfont \AnluTopOne{First}, \AnluTopTwo{second}, and \AnluTopThree{third} place are highlighted per dataset. $^\ast$ preserves the source table\textquotesingle s ranking exclusions for DADA and MOMENT. Excluded cells enter neither average ranks nor counts; a subscript gives the number of ranked datasets when fewer than 10.}\\
\endlastfoot
\rowcolor{AnluTableHead}\multicolumn{15}{@{}l}{\rule{0pt}{12pt}\textbf{VUS-PR$\,\uparrow$}}\\[2pt]
\multicolumn{15}{@{}l}{\color{AnluTableMuted}\scriptsize Zero-shot models}\\[1pt]
TimeRCD & \AnluTopTwo{42.36} & \AnluTopOne{3.09} & 44.06 & \AnluTopThree{96.68} & \AnluTopThree{14.63} & \AnluTopTwo{69.86} & \AnluTopOne{99.27} & \AnluTopTwo{27.48} & 30.50 & \AnluTopTwo{92.21} & 2.40 & 2 & 4 & 2 \\
DADA & {\color{AnluTableMuted}24.66$^\ast$} & {\color{AnluTableMuted}0.57$^\ast$} & 23.17 & 48.24 & 10.61 & 6.42 & 24.42 & 2.94 & \AnluTopThree{33.43} & {\color{AnluTableMuted}83.89$^\ast$} & 8.29$_{7}$ & 0 & 0 & 1 \\
MOMENT & {\color{AnluTableMuted}38.57$^\ast$} & {\color{AnluTableMuted}0.56$^\ast$} & {\color{AnluTableMuted}41.51$^\ast$} & 81.34 & 10.50 & 4.31 & 20.00 & 6.17 & \AnluTopOne{54.61} & {\color{AnluTableMuted}43.39$^\ast$} & 8.83$_{6}$ & 1 & 0 & 0 \\
TimesFM & 18.55 & 0.58 & 21.99 & 36.53 & 10.44 & 6.13 & 29.69 & 6.03 & 21.80 & \AnluTopThree{90.05} & 8.70 & 0 & 0 & 1 \\
Chronos & 18.50 & 0.60 & 21.88 & 33.42 & 10.95 & 8.65 & 27.52 & 6.56 & 19.00 & 85.70 & 7.85 & 0 & 0 & 0 \\
Time-MoE & 16.76 & 0.52 & 20.87 & 23.94 & 9.34 & \AnluTopThree{10.87} & 17.27 & 2.10 & 11.08 & 28.85 & 11.60 & 0 & 0 & 1 \\
MovingVar. & 22.87 & 0.44 & \AnluTopThree{46.53} & 53.53 & 10.70 & 5.17 & 19.83 & 6.90 & 22.85 & 27.65 & 8.50 & 0 & 0 & 1 \\
\multicolumn{15}{@{}l}{\color{AnluTableMuted}\scriptsize Full-shot models}\\[1pt]
TranAD & 20.39 & 0.64 & 24.61 & 89.28 & 13.04 & 5.75 & 14.55 & 2.25 & 12.62 & 35.78 & 8.90 & 0 & 0 & 0 \\
USAD & 16.40 & \AnluTopThree{0.75} & \AnluTopTwo{51.95} & 88.59 & \AnluTopTwo{18.68} & 4.37 & \AnluTopThree{37.21} & 8.85 & 10.07 & 19.45 & 7.10 & 0 & 2 & 2 \\
OmniAnomaly & \AnluTopThree{25.25} & 0.64 & 26.17 & 92.92 & 14.32 & 6.20 & 13.94 & 2.40 & 16.85 & 40.94 & 7.40 & 0 & 0 & 1 \\
LOF & 19.82 & 0.57 & 19.79 & 84.97 & 9.31 & 6.81 & 14.71 & 2.39 & 13.00 & 58.26 & 10.60 & 0 & 0 & 0 \\
IForest & 8.18 & 0.62 & 21.46 & 89.85 & 11.56 & 7.71 & 18.86 & 2.88 & 2.10 & 13.76 & 10.00 & 0 & 0 & 0 \\
Sub-PCA & 23.61 & 0.60 & 43.95 & \AnluTopTwo{97.60} & 10.49 & 3.72 & 29.92 & \AnluTopThree{12.92} & 16.78 & 23.08 & 7.55 & 0 & 1 & 1 \\
DCdetector & 5.18 & 0.59 & 16.49 & 18.36 & 12.32 & 9.37 & 21.97 & 1.53 & 3.26 & 11.63 & 11.70 & 0 & 0 & 0 \\
TFMAE & 4.82 & 0.64 & 16.08 & 20.41 & 11.90 & 9.55 & 18.36 & 2.57 & 5.41 & 38.54 & 10.40 & 0 & 0 & 0 \\
\rowcolor{AnluTableOurs}
\textbf{Anlu} & \AnluTopOne{50.53} & \AnluTopTwo{1.76} & \AnluTopOne{52.56} & \AnluTopOne{99.46} & \AnluTopOne{25.59} & \AnluTopOne{76.38} & \AnluTopTwo{76.49} & \AnluTopOne{56.42} & \AnluTopTwo{43.39} & \AnluTopOne{92.58} & 1.30 & 7 & 3 & 0 \\
\addlinespace[8pt]
\rowcolor{AnluTableHead}\multicolumn{15}{@{}l}{\rule{0pt}{12pt}\textbf{Standard-F1$\,\uparrow$}}\\[2pt]
\multicolumn{15}{@{}l}{\color{AnluTableMuted}\scriptsize Zero-shot models}\\[1pt]
TimeRCD & \AnluTopThree{46.19} & \AnluTopOne{8.63} & 42.57 & \AnluTopThree{92.58} & 23.12 & \AnluTopOne{60.85} & \AnluTopOne{91.97} & \AnluTopTwo{33.47} & \AnluTopThree{41.42} & \AnluTopTwo{89.63} & 2.50 & 3 & 2 & 3 \\
DADA & {\color{AnluTableMuted}32.88$^\ast$} & {\color{AnluTableMuted}0.80$^\ast$} & 25.80 & 51.05 & 15.99 & 2.69 & 19.16 & 3.36 & \AnluTopTwo{45.04} & {\color{AnluTableMuted}86.43$^\ast$} & 10.57$_{7}$ & 0 & 1 & 0 \\
MOMENT & {\color{AnluTableMuted}31.58$^\ast$} & {\color{AnluTableMuted}0.67$^\ast$} & {\color{AnluTableMuted}41.52$^\ast$} & 74.65 & 16.39 & 3.36 & 15.76 & 9.00 & 41.41 & {\color{AnluTableMuted}12.30$^\ast$} & 9.33$_{6}$ & 0 & 0 & 0 \\
TimesFM & 34.57 & 0.83 & 24.56 & 39.12 & 16.73 & 2.96 & 19.84 & 7.86 & 38.45 & \AnluTopThree{85.10} & 8.85 & 0 & 0 & 1 \\
Chronos & 33.13 & 0.99 & 24.33 & 34.49 & 17.47 & 8.74 & 15.41 & 8.21 & 34.47 & 77.78 & 9.10 & 0 & 0 & 0 \\
Time-MoE & 27.67 & 0.45 & 25.05 & 11.46 & 12.16 & \AnluTopThree{17.73} & 19.06 & 4.09 & 20.18 & 36.12 & 11.05 & 0 & 0 & 1 \\
MovingVar. & 23.92 & 0.74 & \AnluTopThree{49.92} & 67.91 & 19.94 & 9.60 & 19.74 & 9.51 & 22.56 & 14.07 & 8.20 & 0 & 0 & 1 \\
\multicolumn{15}{@{}l}{\color{AnluTableMuted}\scriptsize Full-shot models}\\[1pt]
TranAD & 33.28 & 1.46 & 27.42 & 87.22 & 22.36 & 2.63 & 14.44 & 4.40 & 20.40 & 6.32 & 9.40 & 0 & 0 & 0 \\
USAD & 30.55 & \AnluTopThree{3.89} & \AnluTopOne{53.53} & 91.30 & \AnluTopTwo{28.24} & 3.41 & 32.85 & 10.74 & 13.13 & 9.25 & 6.40 & 1 & 1 & 1 \\
OmniAnomaly & \AnluTopTwo{46.85} & 1.44 & 27.79 & 92.25 & \AnluTopThree{23.50} & 0.43 & 13.33 & 5.11 & 29.87 & 30.18 & 7.50 & 0 & 1 & 1 \\
LOF & 32.00 & 1.05 & 23.20 & 80.62 & 12.18 & 4.11 & 15.94 & 4.70 & 22.55 & 62.88 & 9.90 & 0 & 0 & 0 \\
IForest & 8.72 & 0.73 & 27.70 & 90.84 & 19.77 & 3.81 & 20.08 & 4.09 & 2.15 & 2.95 & 10.65 & 0 & 0 & 0 \\
Sub-PCA & 35.74 & 0.83 & 45.09 & \AnluTopTwo{94.80} & 16.05 & 9.56 & 29.04 & \AnluTopThree{15.12} & 25.39 & 11.81 & 6.65 & 0 & 1 & 1 \\
DCdetector & 5.31 & 1.21 & 24.67 & 22.33 & 21.10 & 10.54 & \AnluTopThree{32.91} & 3.18 & 4.63 & 3.97 & 10.30 & 0 & 0 & 1 \\
TFMAE & 9.68 & 0.97 & 24.68 & 19.62 & 20.14 & 11.87 & 20.05 & 2.83 & 15.48 & 29.23 & 10.20 & 0 & 0 & 0 \\
\rowcolor{AnluTableOurs}
\textbf{Anlu} & \AnluTopOne{55.20} & \AnluTopTwo{8.17} & \AnluTopTwo{51.24} & \AnluTopOne{96.04} & \AnluTopOne{32.66} & \AnluTopTwo{58.15} & \AnluTopTwo{67.87} & \AnluTopOne{58.38} & \AnluTopOne{45.05} & \AnluTopOne{89.87} & 1.40 & 6 & 4 & 0 \\
\newpage
\rowcolor{AnluTableHead}\multicolumn{15}{@{}l}{\rule{0pt}{12pt}\textbf{F1-T$\,\uparrow$}}\\[2pt]
\multicolumn{15}{@{}l}{\color{AnluTableMuted}\scriptsize Zero-shot models}\\[1pt]
TimeRCD & 45.40 & \AnluTopOne{9.05} & 49.40 & \AnluTopThree{90.50} & 23.05 & \AnluTopOne{60.59} & \AnluTopOne{89.22} & \AnluTopTwo{41.18} & \AnluTopTwo{44.70} & \AnluTopThree{73.38} & 2.60 & 3 & 2 & 2 \\
DADA & {\color{AnluTableMuted}43.25$^\ast$} & {\color{AnluTableMuted}0.91$^\ast$} & 34.94 & 50.58 & 19.80 & 9.56 & 22.76 & 7.22 & \AnluTopOne{48.47} & {\color{AnluTableMuted}86.15$^\ast$} & 10.00$_{7}$ & 1 & 0 & 0 \\
MOMENT & {\color{AnluTableMuted}34.39$^\ast$} & {\color{AnluTableMuted}0.80$^\ast$} & {\color{AnluTableMuted}49.11$^\ast$} & 74.52 & 19.91 & 9.54 & 26.62 & 13.02 & \AnluTopThree{42.06} & {\color{AnluTableMuted}13.87$^\ast$} & 8.33$_{6}$ & 0 & 0 & 1 \\
TimesFM & \AnluTopTwo{49.29} & 0.93 & 35.13 & 36.49 & 19.80 & 9.58 & 30.20 & 10.78 & 41.58 & \AnluTopOne{84.18} & 7.70 & 1 & 1 & 0 \\
Chronos & \AnluTopThree{46.01} & 1.10 & 34.33 & 33.66 & 19.90 & 13.18 & 27.01 & 10.88 & 39.96 & \AnluTopTwo{78.06} & 7.20 & 0 & 1 & 1 \\
Time-MoE & 26.67 & 0.63 & 36.31 & 12.44 & 19.85 & \AnluTopThree{17.73} & 22.46 & 8.29 & 22.77 & 52.88 & 9.70 & 0 & 0 & 1 \\
MovingVar. & 23.64 & 0.85 & \AnluTopThree{54.93} & 63.78 & 19.93 & 9.59 & 21.68 & 12.85 & 22.02 & 14.73 & 8.90 & 0 & 0 & 1 \\
\multicolumn{15}{@{}l}{\color{AnluTableMuted}\scriptsize Full-shot models}\\[1pt]
TranAD & 22.07 & 1.65 & 36.12 & 82.19 & 22.36 & 9.57 & 20.63 & 7.75 & 21.22 & 7.21 & 9.60 & 0 & 0 & 0 \\
USAD & 21.21 & \AnluTopThree{4.07} & \AnluTopOne{59.20} & 84.28 & \AnluTopTwo{28.23} & 9.54 & \AnluTopThree{39.53} & 14.63 & 14.28 & 9.34 & 6.85 & 1 & 1 & 2 \\
OmniAnomaly & \AnluTopOne{53.04} & 1.61 & 37.79 & 87.37 & \AnluTopThree{23.48} & 9.68 & 20.98 & 8.47 & 35.41 & 30.79 & 6.40 & 1 & 0 & 1 \\
LOF & 29.48 & 1.15 & 33.37 & 75.13 & 19.80 & 9.60 & 21.42 & 8.31 & 24.28 & 68.87 & 9.10 & 0 & 0 & 0 \\
IForest & 7.87 & 0.84 & 19.76 & 83.04 & 0.00 & 9.54 & 12.07 & 6.36 & 4.45 & 2.60 & 13.85 & 0 & 0 & 0 \\
Sub-PCA & 34.46 & 0.98 & 51.13 & \AnluTopTwo{94.80} & 20.30 & 9.54 & 29.72 & \AnluTopThree{18.90} & 25.31 & 12.30 & 7.05 & 0 & 1 & 1 \\
DCdetector & 6.77 & 1.32 & 32.25 & 22.44 & 21.13 & 10.53 & 33.66 & 6.62 & 7.44 & 4.85 & 10.50 & 0 & 0 & 0 \\
TFMAE & 20.13 & 1.07 & 32.00 & 19.45 & 20.18 & 11.82 & 21.25 & 5.90 & 19.15 & 31.00 & 11.20 & 0 & 0 & 0 \\
\rowcolor{AnluTableOurs}
\textbf{Anlu} & 44.69 & \AnluTopTwo{7.29} & \AnluTopTwo{57.84} & \AnluTopOne{95.44} & \AnluTopOne{33.09} & \AnluTopTwo{57.59} & \AnluTopTwo{68.65} & \AnluTopOne{58.04} & \AnluTopThree{42.06} & 64.70 & 2.45 & 3 & 4 & 1 \\
\addlinespace[8pt]
\rowcolor{AnluTableHead}\multicolumn{15}{@{}l}{\rule{0pt}{12pt}\textbf{Affiliation-F$\,\uparrow$}}\\[2pt]
\multicolumn{15}{@{}l}{\color{AnluTableMuted}\scriptsize Zero-shot models}\\[1pt]
TimeRCD & \AnluTopThree{87.34} & \AnluTopOne{74.54} & \AnluTopTwo{93.60} & \AnluTopThree{99.66} & \AnluTopOne{85.02} & \AnluTopOne{88.99} & \AnluTopOne{99.04} & \AnluTopTwo{84.34} & 90.74 & \AnluTopOne{98.38} & 2.00 & 5 & 2 & 2 \\
DADA & {\color{AnluTableMuted}90.13$^\ast$} & {\color{AnluTableMuted}67.66$^\ast$} & 85.71 & 97.35 & 69.79 & 65.18 & 82.17 & 72.21 & \AnluTopTwo{93.74} & {\color{AnluTableMuted}96.02$^\ast$} & 8.57$_{7}$ & 0 & 1 & 0 \\
MOMENT & {\color{AnluTableMuted}88.67$^\ast$} & {\color{AnluTableMuted}66.76$^\ast$} & {\color{AnluTableMuted}89.74$^\ast$} & 94.94 & 75.97 & 59.13 & 62.07 & 75.77 & \AnluTopOne{95.22} & {\color{AnluTableMuted}86.71$^\ast$} & 7.83$_{6}$ & 1 & 0 & 0 \\
TimesFM & 81.86 & 66.95 & 78.58 & 93.30 & 69.88 & 67.14 & 86.02 & 70.03 & 79.00 & 91.79 & 9.30 & 0 & 0 & 0 \\
Chronos & \AnluTopTwo{90.41} & 67.89 & 85.92 & 97.44 & 69.72 & 67.89 & \AnluTopThree{93.27} & 74.35 & 90.79 & \AnluTopThree{97.39} & 5.90 & 0 & 1 & 2 \\
Time-MoE & 77.02 & 67.23 & 79.33 & 86.64 & 71.19 & 60.98 & 58.58 & 73.56 & 80.24 & 75.77 & 10.60 & 0 & 0 & 0 \\
MovingVar. & 84.93 & 67.56 & 91.06 & 89.91 & 74.77 & 67.14 & 82.97 & \AnluTopThree{78.44} & 89.25 & 84.22 & 6.10 & 0 & 0 & 1 \\
\multicolumn{15}{@{}l}{\color{AnluTableMuted}\scriptsize Full-shot models}\\[1pt]
TranAD & 83.63 & 67.28 & 89.65 & 99.33 & 71.56 & 61.03 & 59.07 & 73.31 & 84.91 & 83.78 & 8.30 & 0 & 0 & 0 \\
USAD & 71.39 & 67.81 & 90.96 & 85.80 & 76.48 & 55.60 & 75.81 & 76.00 & 65.70 & 60.16 & 9.90 & 0 & 0 & 0 \\
OmniAnomaly & 81.05 & 67.35 & \AnluTopThree{91.83} & \AnluTopTwo{99.76} & \AnluTopThree{78.16} & 61.26 & 56.99 & 73.53 & 78.96 & 79.76 & 8.25 & 0 & 1 & 2 \\
LOF & 82.34 & \AnluTopThree{68.44} & 73.97 & 99.31 & 66.76 & 63.85 & 58.42 & 73.53 & 81.28 & 84.38 & 8.95 & 0 & 0 & 1 \\
IForest & 52.75 & \AnluTopTwo{68.82} & 39.17 & 85.78 & 0.00 & 70.09 & 41.83 & 50.56 & 42.66 & 31.14 & 12.80 & 0 & 1 & 0 \\
Sub-PCA & 75.61 & 66.90 & 88.51 & 99.03 & 71.37 & 67.14 & 83.80 & 76.66 & 76.72 & 75.18 & 8.60 & 0 & 0 & 0 \\
DCdetector & 72.05 & 67.91 & 71.40 & 73.57 & 69.75 & \AnluTopThree{72.20} & 73.84 & 70.18 & 72.73 & 70.42 & 11.10 & 0 & 0 & 1 \\
TFMAE & 78.60 & 67.50 & 75.32 & 86.19 & 70.30 & 68.17 & 69.62 & 70.60 & 79.92 & 84.10 & 9.70 & 0 & 0 & 0 \\
\rowcolor{AnluTableOurs}
\textbf{Anlu} & \AnluTopOne{93.31} & 65.35 & \AnluTopOne{94.34} & \AnluTopOne{99.80} & \AnluTopTwo{84.25} & \AnluTopTwo{88.83} & \AnluTopTwo{93.62} & \AnluTopOne{93.54} & \AnluTopThree{92.48} & \AnluTopTwo{98.25} & 2.90 & 4 & 4 & 1 \\
\end{longtable}
\endgroup
\clearpage

\subsection{Results and reference intervention}
\label{sec:benchmark-result}

Table~\ref{tab:reference-results} compares Anlu with the frozen TSFM and with the zero-reference condition. Anlu attains a mean VUS-PR of 0.607 with the original reference prefixes, 0.065 above the frozen TSFM at 0.542. Setting the reference to zero lowers Anlu's mean to 0.499. Relative to the zero-reference condition, the original prefix has the higher score on 248 sequences, the lower score on 82, and the same score on 20.

\begin{table}[!htbp]
\centering
\caption{Mean VUS-PR on all 350 TSB-AD-U Eval sequences under the protocol in Section~\ref{sec:evaluation-setup}.}
\label{tab:reference-results}
\begin{tabularx}{0.88\linewidth}{@{}Yr@{}}
\toprule
Model and reference condition & Mean VUS-PR \\
\midrule
Frozen TSFM, no reference pathway & 0.542 \\
Anlu, reference set to zero after normalization & 0.499 \\
Anlu, original reference prefix & \textbf{0.607} \\
\bottomrule
\end{tabularx}
\end{table}

The two Anlu conditions share weights, queries, reference lengths, and validity masks and differ only in the normalized reference values.

\begin{figure}[!htbp]
 \centering
 \anlufigure[0.70\linewidth]{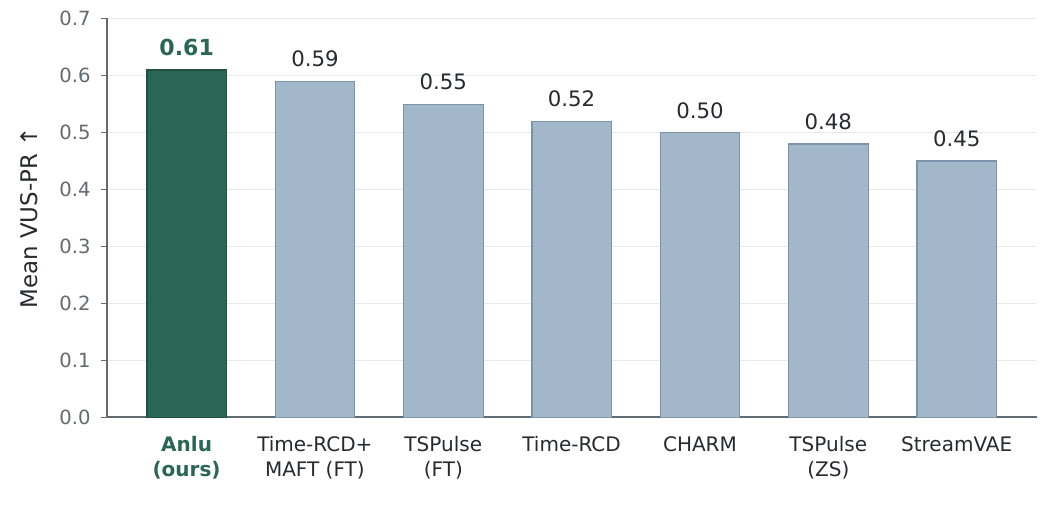}
 \caption{Mean VUS-PR on TSB-AD-U. The other six values come from the official leaderboard snapshot \citep{tsbadleaderboard2026}. FT and ZS follow that snapshot's fine-tuned and zero-shot labels. Anlu uses the protocol in Section~\ref{sec:evaluation-setup}.}
 \label{fig:benchmark}
\end{figure}

Figure~\ref{fig:benchmark} plots Anlu's mean VUS-PR beside six entries from the TSB-AD-U leaderboard snapshot retrieved on October 3, 2026 \citep{tsbadleaderboard2026}.

\subsection{Score profiles}
\label{sec:qualitative}

\begin{figure}[!htbp]
 \centering
 \anlufigure[\linewidth]{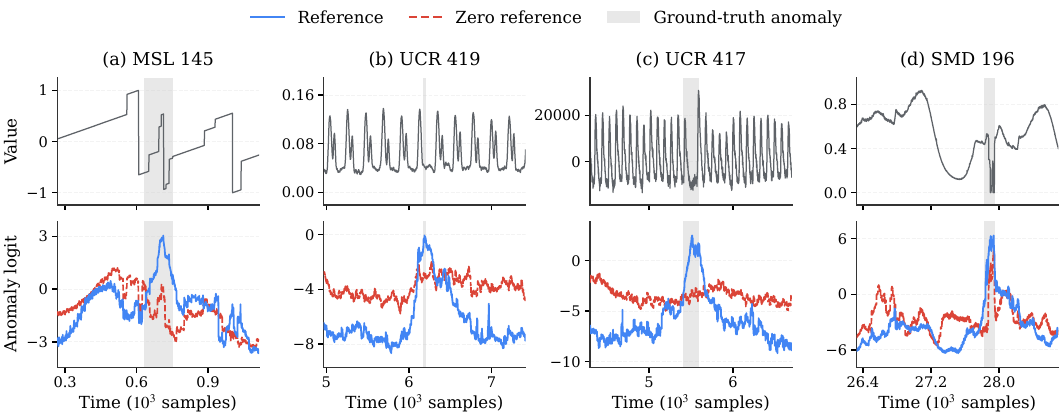}
 \caption{Queries and anomaly logits for MSL 145, UCR 419, UCR 417, and SMD 196. The top row is the query signal. The bottom row is the anomaly logit $u_1-u_0$. Solid blue curves use the original reference prefix. Dashed red curves set that prefix to zero after normalization. Gray bands mark annotated anomalies. The query, validity mask, and reference length are shared. Horizontal axes give sample indices in thousands.}
 \label{fig:reference-intervention}
\end{figure}

Figure~\ref{fig:reference-intervention} shows MSL 145, UCR 419, UCR 417, and SMD 196. The curves are merged anomaly logits from Anlu. None of the four reference prefixes contains an annotated anomaly. Appendix~\ref{app:reference-cases} lists the sequences, the display windows, and the selection rule.

\section{Conclusion}
\label{sec:conclusion}

Anlu freezes a pretrained anomaly detector and trains a reference memory with gated adapters at four encoder depths. For the synthetic tasks, one stored query is paired with two references, and each label follows the normal behavior of the accompanying reference.

On all 350 TSB-AD-U Eval sequences, Anlu attains a mean VUS-PR of 0.607 with the dataset reference prefixes, compared with 0.542 for the frozen TSFM. The zero-reference condition lowers Anlu's mean to 0.499.

\begingroup
\small
\raggedright
\setlength{\bibsep}{3pt}
\bibliographystyle{plainnat}
\bibliography{main}
\endgroup

\appendix
\section{Conditional Rule Specification}
\label{app:generation}

\rev{Table~\ref{tab:rules} summarizes the six process-rule families. Within each family, waveform shape, period, phase, noise, and signal scale vary randomly. Each candidate unit renders one query and two references. A unit is accepted only if each reference repeats its rule often enough to support a judgment, the difference between the two rules is visible in the rendered waveform, the labels agree with the rules, and at least one query position changes its label.}

\begin{table}[!htbp]
\centering
\small
\caption{Process-rule families and their conditional targets.}
\label{tab:rules}
\begin{tabularx}{\linewidth}{@{}L{0.19\linewidth}YY@{}}
\toprule
Rule family & Normal-reference evidence & Conditional query judgment \\
\midrule
Transition speed & Repeated transitions between state levels & Whether transition speed lies within the supported range \\
Skipped states & Adjacent or multi-step state transitions & Whether a transition skips more states than the supported limit \\
Attenuation or absence & Event strengths and repeated admissible omissions & Whether attenuation or omission is allowed at the relevant slot \\
Event enhancement & Ordinary and enhanced event strengths & Whether enhancement exceeds the supported envelope \\
Timing shift & Event offsets relative to the process cycle & Whether timing lies outside the supported range \\
Idle-interval events & Repeated occupancy or absence at a scheduled slot & Whether an added event is admissible at that slot \\
\bottomrule
\end{tabularx}
\end{table}

The event-strength rule uses an envelope of realized nonnegative strengths $v_1,\ldots,v_n$ at a scheduled event slot:
\begin{equation}
 I_r=\left[\max\!\left(0,\min_jv_j-\delta\right),\max_jv_j+\delta\right],
 \qquad\delta=0.14.
 \label{eq:strength-envelope}
\end{equation}
\rev{The tolerance $\delta$ is measured in the generator's event-strength units, before the normalization in Equation~\eqref{eq:normalization}. An absence counts as normal only if the reference shows at least two absences at that slot; an envelope that reaches zero does not by itself make an absence normal. The generator derives labels from the known schedule and the rendered waveform, and the detector receives neither.}

\section{Implementation and Evaluation Details}
\label{app:implementation}

\subsection{Data and training}
\label{app:training}

\rev{Each generation unit produces either a reference-conditioned pair, with probability 0.5, or a single anomaly-injection example. Because a pair contributes two labeled examples, about two thirds of the training examples are reference-conditioned. An anomaly-injection example contains a query with injected anomalies and no reference. It is paired with an all-zero raw record of the query length, which becomes a constant series after the query-based normalization in Equation~\eqref{eq:normalization}.}

\rev{Table~\ref{tab:config} lists the model and optimization settings.}

\begin{table}[!htbp]
\centering
\small
\caption{Anlu configuration. Encoder block indices start at 1.}
\label{tab:config}
\begin{tabularx}{\linewidth}{@{}L{0.30\linewidth}Y@{}}
\toprule
Component & Setting \\
\midrule
Input and encoders & Univariate input, $C=1$. Separate query and reference encoders with the same initialization \\
Encoder architecture & 8 blocks, width 512, 8 heads, feed-forward width 2,048 \\
Patching and projection & Patch size 16, pointwise feature width 256 \\
Position information & Encoder rotary embeddings, no added resampler position embeddings \\
Reference memory & 32 tokens per channel, 1 resampling layer \\
Adapters & After blocks 3, 5, 7, and 8, with both scalar gates initialized to zero \\
Frozen parameters & Both encoders, pointwise projection, detection head, reconstruction head \\
Trainable parameters & Reference resampler and gated adapters, including gates \\
Loss & Pointwise binary cross-entropy over valid positions plus masked-query MSE, with equal weights \\
Normalization and crops & Full-query statistics shared with the reference, query crops up to 10,000 samples, reference crops of 90\%--100\% \\
Optimizer & AdamW, constant learning rate $10^{-4}$, weight decay $10^{-5}$, batch size 24 \\
Regularization & Dropout 0.1, independent 15\% patch masks on the query and the reference \\
\bottomrule
\end{tabularx}
\end{table}

\subsection{Evaluation implementation}
\label{app:evaluation}

\rev{We use the official TSB-AD-U evaluation split of 350 univariate sequences. Reference prefixes range from 500 to 10,000 samples. Queries longer than 10,000 samples are scored in windows of 10,000 samples with stride 1,500, with a final window aligned to the end of the sequence, and overlapping anomaly logits are averaged.}

\rev{VUS-PR is computed from the merged anomaly logits with the TSB-AD implementation and 250 score thresholds. Following TSB-AD, the maximum temporal tolerance of each sequence is set from its estimated dominant period.}

\section{Reference-Effect Cases}
\label{app:reference-cases}

\rev{Figure~\ref{fig:reference-intervention} shows four sequences on which the original-prefix condition has a higher VUS-PR than the zero-reference condition and the reference prefix contains no annotated anomaly. Each display window is centered on the annotated anomaly with the largest logit change between the two conditions, measured by the mean within-sequence percentile rank. Table~\ref{tab:cases} lists the sequences. The curves are computed with the protocol in Section~\ref{sec:evaluation-setup} and then cropped for display.}

\begin{table}[!htbp]
\centering
\small
\caption{Sequences shown in Figure~\ref{fig:reference-intervention}, from left to right. Display ranges are zero-based sample indices after rows with missing values are removed.}
\label{tab:cases}
\begin{tabular}{@{}lllrr@{}}
\toprule
Panel & Source & TSB-AD index & Reference length & Display range \\
\midrule
1 & MSL & 145 & 530 & 267--1,114 \\
2 & UCR & 419 & 3,000 & 4,968--7,412 \\
3 & UCR & 417 & 2,750 & 4,300--6,700 \\
4 & SMD & 196 & 7,174 & 26,325--28,695 \\
\bottomrule
\end{tabular}
\end{table}

\end{document}